%% file: main.tex
\documentclass{styles/svproc}

\usepackage[utf8]{inputenc}
\usepackage{booktabs}
\input{Imports/librairies.tex}
\input{Imports/commands.tex}
\input{Imports/notations.tex}


\newcommand{\repeatthanks}{\textsuperscript{\thefootnote}}

\begin{document} 
\title{Extended URDF: Accounting for parallel mechanism in robot description}
\author{Virgile Batto\thanks{Equal contribution. Authors are listed in alphabetical order.}\inst{1, 2} \and Ludovic De Matteïs\repeatthanks\inst{1, 3} \and Nicolas Mansard\inst{1, 4}}
\authorrunning{Batto, De Matteïs, et al.}

\institute{
LAAS-CNRS, 7 Avenue du Colonel Roche, 31400, Toulouse, France
\and
Inria Bordeaux, 200 Avenue de la Vieille Tour, 33405, Talence, France
\and
Inria - Département d’Informatique de l'Ecole normale supérieure, PSL
Research University, Paris, France
\and
ANITI, 41 allées Jules Guesde, 31000, Toulouse, France
}

\date{January 2025}
\maketitle





\input{tex/0.Abstract}
\input{tex/1.Introduction}
\input{tex/2.background}

\input{tex/3.ExtendingDescriptionFormat}
\input{tex/4.Test}

\input{tex/5.Conclusion}
\bibliographystyle{styles/bibtex/spmpsci}
\bibliography{references}

\end{document}

%% file: Imports/librairies.tex
\usepackage{graphicx}
\usepackage{xcolor}
\definecolor{mygreen}{rgb}{0,0.6,0}
\definecolor{mygray}{rgb}{0.5,0.5,0.5}
\definecolor{mymauve}{rgb}{0.58,0,0.82}

\usepackage{mathtools}
\usepackage{amssymb}
\usepackage[bottom]{footmisc}
\usepackage{hyperref}

\usepackage{url}

\usepackage{subcaption}

\usepackage{mathtools}

\usepackage{fourier} 
\usepackage{array}
\usepackage{makecell}


%% file: Imports/commands.tex
\definecolor{ld_blue}{rgb}{0, 0, 0.6}
\definecolor{el_purple}{rgb}{0.6, 0, 0.6}
\definecolor{nm_orange}{rgb}{0.8, 0.4, 0.0}
\definecolor{jc_pink}{rgb}{0.6, 0.0, 0.3}

\definecolor{deepblue}{rgb}{0,0,0.5}
\definecolor{deepred}{rgb}{0.6,0,0}
\definecolor{deepgreen}{rgb}{0,0.5,0}
\definecolor{lightgray}{gray}{0.95} 

%% file: Imports/notations.tex
\newcommand{\kron}{\otimes}

\newcommand{\SE}[1]{\mathrm{SE}(#1)}

\newcommand{\matr}[1]{\boldsymbol{#1}}

\newcommand{\zeros}[2]{
\ifthenelse{\equal{#2}{1}}{\vect{0}_{#1}}{\matr{\cancel{O}}_{#1 \times #2}}
}
\newcommand{\ones}[2]{
\ifthenelse{\equal{#2}{1}}{\vect{1}_{#1}}{\matr{1}_{#1 \kron #2}}
}

\newcounter{simulationcase}

\newcommand{\old}[1]{}  



%% file: tex/0.Abstract.tex
\begin{abstract}
Robotic designs played an important role in recent advances by providing powerful robots with complex mechanics. Many recent systems rely on parallel actuation to provide lighter limbs and allow more complex motion. However, these emerging architectures fall outside the scope of most used description formats, leading to difficulties when designing, storing, and sharing the models of these systems. This paper introduces an extension to the widely used Unified Robot Description Format (URDF) to support closed-loop kinematic structures. Our approach relies on augmenting URDF with minimal additional information to allow more efficient modeling of complex robotic systems while maintaining compatibility with existing design and simulation frameworks. This method sets the basic requirement for a description format to handle parallel mechanisms efficiently. We demonstrate the applicability of our approach by providing an open-source collection of parallel robots, along with tools for generating and parsing this extended description format. The proposed extension simplifies robot modeling, reduces redundancy, and improves usability for advanced robotic applications.

\keywords{Robotics, Modeling, Description Format, Humanoids, Parallel Robots, Kinematics, Standardization}
\end{abstract}

%% file: tex/1.Introduction.tex
\section{Introduction}

The democratization of electric actuators has led to large usage of serial robots with low-complexity mechanics but a high number of degrees of freedom (DoF).
Along with methods and tools to simulate the robots\cite{justincarpentierPinocchioLibraryFast2019}\cite{todorov2012mujoco}, description formats were developed, allowing researchers and manufacturers to share their models.
One of the most famous format is the Unified Robot Description Format (URDF) \cite{quigley2015programming} that remains to date the basic standard for robot modeling \cite{tola2023understanding}.
Yet, URDF only applies to serial mechanisms.
While parallel kinematics has long been recognized for dedicated applications in manipulation or haptic, it now becomes increasingly important through hybrid architectures in legged robotics. 
Most recent bipedal platforms employ parallel mechanisms, especially for the ankles actuation, as seen in robots like Digit \cite{digit}, Tello \cite{sim2022tello}, Tesla Optimus \cite{TeslaOptimus} and Unitree G1 \cite{HumanoidRobotG1_Humanoid}.
These architectures reduce the foot effective inertia and enhance the robot dynamic capabilities \cite{virgilebattoComparativeMetricsAdvanced2023}.
So far, the community is lacking of efficient tools to accurately model and control these systems. Consequently, a part of the kinematics is often neglected for simulation and control of these robots, leading to suboptimal behavior \cite{dematteis:hal-04716938}.
Even though simulation frameworks have begun integrating algorithms to handle such architectures \cite{justincarpentierProximalSparseResolution2021,royfeatherstoneRigidBodyDynamics2007}, there exists no robot description format suitable for parallel mechanisms with universal acceptation.
Recent solutions like URDF+ \cite{chignoli2024urdf+} and SDF \cite{SDF} all suffer of significant conceptual flaws, as we will discuss with more details in Sec.~\ref{subsec:description_formats}.

In this paper, we argue that a description format able to handle parallel mechanisms should be based on an underlying serial kinematic chain and make closure constraints explicit.
We propose a complete solution by extending URDF with additional information describing the closures.
The proposed format is generic and remains flexible for user-specific applications.
To illustrate its use, we present a collection of parallel robots modeled with our approach, using the Pinocchio library \cite{justincarpentierPinocchioLibraryFast2019}.

The paper is organized as follows. In Sec.~\ref{sec:background} we provide background on robot modeling and closed-loop constraints. In section Sec.~\ref{sec:extending_urdf} we describe how we extended the URDF to handle a wider variety of architectures. Section Sec.~\ref{sec:toolbox} shows examples of robots modeled with our approach. Finally, Sec.~\ref{sec:conclusion} concludes the work and discuss future work.

%% file: tex/2.background.tex
\section{Background}\label{sec:background}
\subsection{Multi-Body Systems} \label{subsec:multibody_system}
Modeling a multi-body system is typically achieved using graphs that represent the system's kinematic tree. For robotic systems, this connectivity graph is undirected and connected, with nodes representing bodies with inertial properties and edges corresponding to joints. Some robot architectures introduce loops in this graph, indicating kinematic closures. To accurately model robotics systems, one must define the position and velocity of each joint using so-called generalized coordinates.
In this setting, the system configuration can be expressed by a single vector $q$ of size $n_q$ defining the position of each joint and the system articular velocity is given by the vector $\dot q$ of size $n_v$, where in general for legged robotics $n_q \neq n_v$.
We describe the transform from a frame A to a frame B by an element of $\SE3$ denoted $^AM_B$.
%
\subsection{Loop Constraints} \label{subsec:loop_constraints}
The kinematic closure can be modeled by adding a bilateral motion constraint to the system, fixing the relative placement of two contact points. 
This constraint appears in the equation of motion as an force applied on the system joints. 
This leads to the following procedure; 1) Extract a representative spanning tree of the closed-loop system; 2) Add constraints to the equation of motion to account for kinematic closure; 3) Solve simultaneously these equations for the system acceleration and contact forces. \cite{justincarpentierProximalSparseResolution2021}
These closed-loop constraints can be expressed either in explicit or implicit form.
The explicit formulation express a relation between an independent subset of configuration variables and the complete system configuration, that can be formulated at the position, velocity or acceleration level \cite{royfeatherstoneRigidBodyDynamics2007}
This expression is in general difficult to obtain is more likely hard to transpose to different systems.\\
In the implicit formulation, the constraint takes the form
\begin{equation}
    \phi(q) = 0; \qquad K(q)\dot{q} = 0; \qquad K(q)\ddot{q} = k(q, \dot{q})
\end{equation}
which has been shown to be general, with existing methods to solve the corresponding equations of motion, with $K = \frac{\partial \Phi}{\partial q}$ the constraint Jacobian. \cite{justincarpentierProximalSparseResolution2021}.
In this paper we will focus on generating, storing and sharing efficiently the data required to formulate these constraints and solve the corresponding equation of motion. 

\subsection{Existing Description Formats} \label{subsec:description_formats}
The models of robotics systems are usually contained in description files, that allows describing the structure of the robot along with other information such as frames of interest and joint limits. 
To the best of our knowledge, three main description formats are in use nowadays, URDF, SDF and the more recent MCJF.
The first two have been used for a long time in the robotics community and now come with efficient file generators and parsers. 
However, they struggles to handle robots with closed kinematic loops for different reasons.
On one hand, the URDF relies on the tree-like structure of a robot model and is limited in joint type, as it can only handle revolute, linear, prismatic, planar, and floating joints, missing the now widely used ball joints. 
On the other hand, the SDF handles all joint types and is not restricted to tree-like structure, thus allowing representing closed-loop systems. 
However, it lacks information to efficiently parse and use parallel mechanisms.
For instance the inertia properties repartition in closed-loops, the constraint placements, and the constraint formulation are not clearly defined, yielding unrealistic splittings.
It also tends to generate a null inertia link near the tree extremity, yielding an generalized inertia matrix that lacks strict positivity.
The MCJF and srobot (format used on the Drake\cite{drake} simulator) format on the opposite, are designed to be able to work with kinematic closures, but stays for now restricted to the MuJoCo/Drake simulator. Therefore, using this format would require users to write custom and complex file parsers and generators for their applications. 
Thus, as highlighted on the table \ref{tab:robot_formats}, there is no existing format or method to correctly handle systems with kinematic closures while maintaining the ease of use required in most applications. 

\begin{table}[h]
    \centering
    \small 
    \renewcommand{\arraystretch}{1.2} 
    \setlength{\tabcolsep}{4pt} 
    \begin{tabular}{p{2.5cm}|c|c|c|p{2.5cm}} 
        \toprule
        \thead{Format} & \thead{Ease to \\ generate} & \thead{Models \\closed-kinematics} & \thead{Models \\ the actuation} & \thead{Restrictions} \\
        \midrule
        URDF          & Easy     & No        & No   & General Use  \\
        SDF           & Moderate & Partially   & No   & Gazebo Only  \\
        URDF+         & Hard     & Yes       & No   & Custom Implementations \\
        MJCF (MuJoCo)         & Moderate & Yes       & Yes  & MuJoCo Only  \\
        Srobot (Drake)& Hard     & Yes       & Yes  & Drake Only  \\
        Extended URDF (ours) & Easy     & Yes       & Yes  & Pinocchio - Easily adaptable \\
        \bottomrule
    \end{tabular}
    \caption{Comparison of robot description formats and their main caracteristics}
    \label{tab:robot_formats}
\end{table}

\subsection{Closed-Loop Algorithms} \label{subsec:closed_loop_algorithms}
In order to use closed-loop algorithms, one should be able to define implicit constraints by simply parsing the description files \cite{justincarpentierProximalSparseResolution2021}.
To do so, several information is required, that is the constraint type - i.e. what form the implicit constraint should take -, the constraint placement, usually defined by a parent joint and a relative placement with respect to it and the underlying serial model to use.
With this information, one can write the equation of motion of the serial model and add bilateral contact constraints.
In this paper, we will limit the discussion to two types of constraints.
The 6D constraint imposes the contact frames to have the same position and same orientation
\begin{equation}
    \log_{SE3}\left(^AM_B(q)\right) = 0
\end{equation}
where A and B denote the contact frames on both side of the constraint (cf Figure~\ref{fig:digit_6d_3d}).
The 3D constraint imposes the contact points to have the same position but do not constrain any orientation, leaving
\begin{equation}
    \vec{AB}(q) = 0
\end{equation}
While our description method mostly targets these two formulations, we let it general enough to handle new constraint types.

%% file: tex/3.ExtendingDescriptionFormat.tex
\section{Extension of Description Format} \label{sec:extending_urdf}
\subsection{Method Overview} \label{subsec:method_overview}
As discussed in Sec.~\ref{sec:background}, we propose a method to overcome limitations in current description formats. We propose to easily generate the description files from a CAD software while being adaptable for different architectures, constraints, and joint types. Additionally, we found information on the robot actuation to be essential for control problems.

We extend the well-defined URDF format, preserving its advantages and widespread adoption and extend it using a single YAML file, which stores structured information efficiently while allowing both automated generation and manual modifications. The simplicity of YAML enables users to add custom data, such as frames of interest or velocity limits, and easily adapt parsers and generators.
By using an external file, this approach ensures backward compatibility when only URDF is required. To fully support kinematic closures, the extended format must overcome URDF’s limitations by incorporating closed-loop information, defining system actuation, and supporting additional joint types beyond the URDF standard.
\subsection{Handling Kinematic Closures} \label{subsec:handling_kinematic_closures}
As mentioned in section~\ref{subsec:loop_constraints}, kinematic closures are usually written as a constraint on points relative placements.
While considering 6D constraint only may be sufficient for modeling kinematic closures, handling several constraint types can be beneficial in certain context.
For instance, if one parent joint of the 6D constraint is a ball joint, one can rewrite the constraint using a 3D formulation in place of the ball joint.
This yields the same kinematic behavior while maintaining a smaller system state.
\begin{figure}
    \centering
    \includegraphics[width=0.6\linewidth]{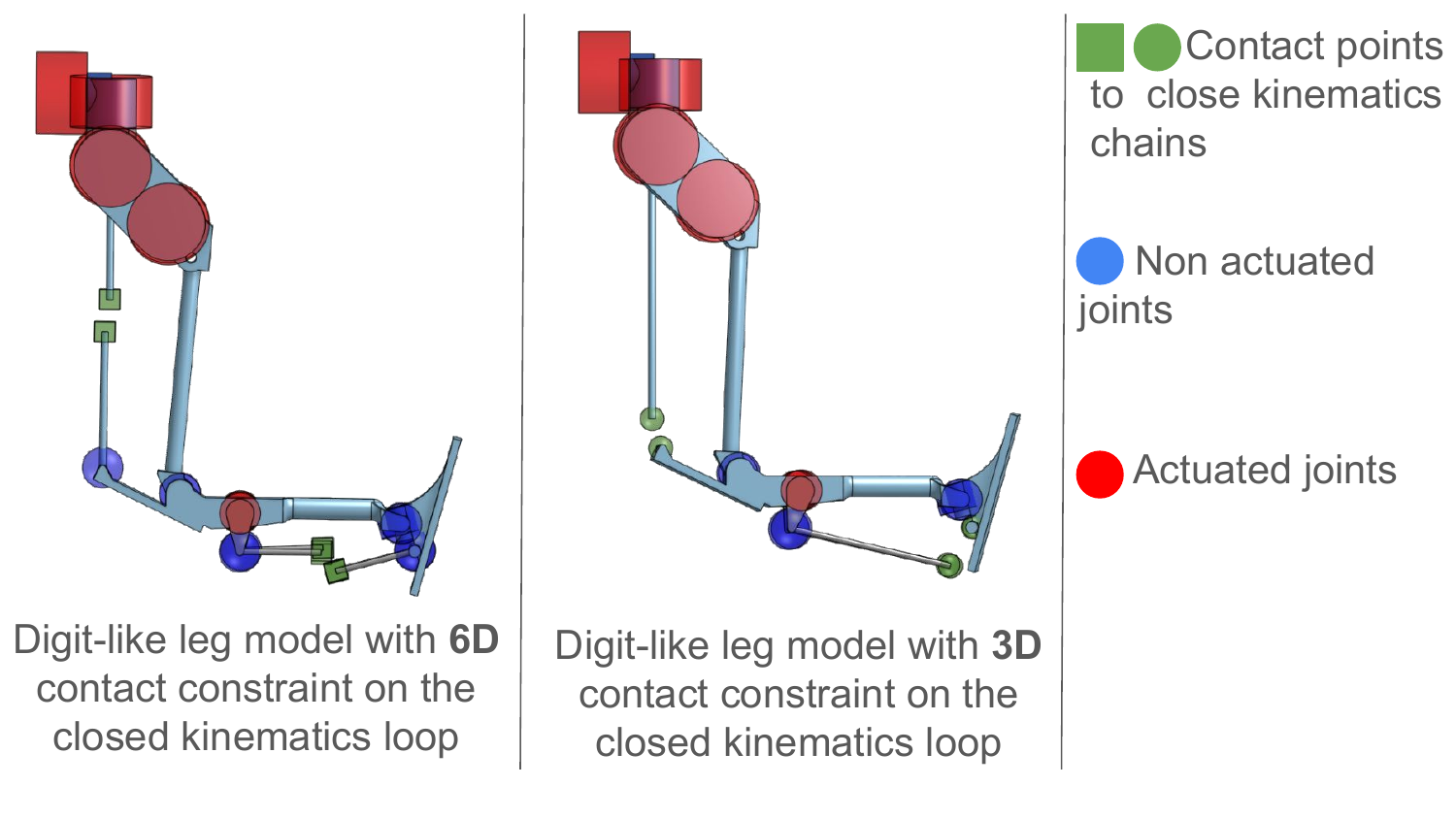}
    \caption{Information needed to fully exploit a parallel mechanism leg, with either 6d or 3d contact}
    \label{fig:digit_6d_3d}
\end{figure}
Fig~\ref{fig:digit_6d_3d} presents a model of a parallel robot - based on the Digit architecture \cite{digit} - modeled with either 6D or 3D constraints.\\

To store the kinematic closure information in a description file, we need to specify the type of constraint (3D or 6D), the parent joints of the contact points, and their placements with respect to their parents.
%
The two last information can be replaced by the knowledge of two frames associated with the contact points, which can in turned be described in the URDF, yielding a lighter YAML extension.
Therefore, we propose to dedicate two fields in the extension file for a) defining the type of constraint for each kinematic closure and b) point to the contact frames - which are defined in the URDF file.
In order to ensure that the inertia of each link is correct, we model the underlying serial model in a CAD software, with added contact frame, before exporting to URDF.
\subsection{Description of the Under-actuation} \label{subsec:actuation_information}
By definition, parallel mechanisms must have actuated and non-actuated joints. 
For instance, on a leg of the Digit robot, there are 6 actuated revolute joints, 3 passive revolute joint, and 6 free spherical joints as shown in figure \ref{fig:digit_6d_3d}.
Note that using 3D contacts can help reducing the dimension of the problem by transforming a ball joint into a single 3D constraint.\\
The knowledge of actuated joints is mandatory for many problem such as control.
Hence the creation of another field in the extension file for storing actuation information.
As there are often more passive joints than actuated ones, we choose to include the names of the actuated joints in the YAML file.
The rest of the joint information (limits, type...) is stored classically in the URDF.

\subsection{Extension to Specific Joints} \label{subsec:joints_extension}

To overcome URDF's limitations in joint type representation, we use the YAML file to specify joint replacements, enabling support for additional joint types. During parsing, this allows the automatic substitution of one or more joints with alternative joints, which is crucial for kinematic closures that typically involve ball joints or U-joints, both unsupported in URDF. Additionally, our implementation detects three consecutive, concurrent revolute joints without intermediate link inertia and automatically replaces them with a spherical joint, improving efficiency and accuracy in modeling.
We also believe that future implementation of URDF will nativelly solve this issue.

\subsection{Parsing the Extended Files} \label{subsec:parsing}


In order to use the extended format in different applications, we propose a method to easily load and generate it. Our implementation includes an algorithm that loads URDF files into the Pinocchio library \cite{justincarpentierPinocchioLibraryFast2019}, extending them with closure and actuation details. When parsing the YAML file, we enforce joint type modifications, create an actuation model, and generate constraint models for constrained dynamics algorithms. The YAML format's readability and existing parsing tools facilitate replication across different simulators.
For model generation, URDF files can be created using CAD-embedded tools \cite{onshape2robot}\cite{solidworks2robot}. 
We also provide a script that automatically generates the YAML file based on user-defined naming conventions, while keeping it simple enough for manual editing when needed.

%% file: tex/4.Test.tex
\section{Examples of Robots Models} \label{sec:toolbox}
To showcase the method’s versatility, we model robots with closed kinematic chains using our extended URDF format. The URDFs are generated via Onshape to robot \cite{onshape2robot}, with YAML files written manually or following a naming convention. Below, we present models of Digit and Kangaroo.

\subsection{Digit}
\begin{figure}[t!]
     \centering
     \begin{subfigure}[b]{0.45\textwidth}
         \centering
         \includegraphics[width=0.25\textwidth, trim={3cm 1cm 2cm 0cm}, clip]{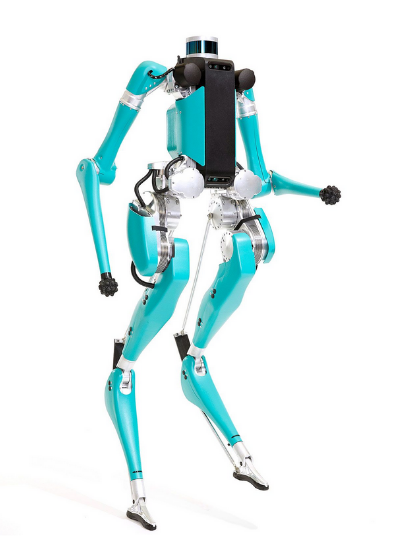}
         \caption{Digit Robot}
         \label{fig:digitRobot}
     \end{subfigure}
     \hspace{0cm}
     \begin{subfigure}[b]{0.45\textwidth}
         \centering
         \includegraphics[width=0.25\textwidth, trim={3cm 1.5cm 3cm 5cm}, clip]{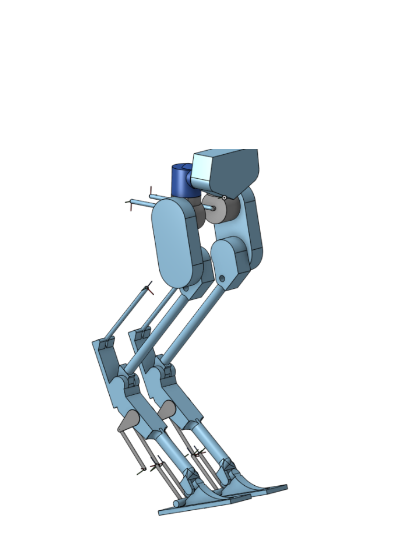}
         \caption{Modeled digit robot}
         \label{fig:digitModel}
     \end{subfigure}
    \caption{Digit real and modeled robot. In the model, the 6D contacts constraints are "opened" for better visualization}
    \label{fig:digit}
\end{figure}

Digit is a humanoid robot designed for factory work, carrying boxes while walking. We model only its legs, as the upper body is purely serial and compatible with standard URDF. Fig.~\ref{fig:digit} shows the real and modeled robots. 
The real and modeled robots are presented in Fig.~\ref{fig:digit}.
Each leg has 6 motors for 6 DoF, with three kinematic closures per leg—one for knee motion via a hip motor and two for ankle motion via calf motors. Our model consists of a 27 DoF serial structure with 18 constrained DoF ($3 \times 6D$ contact constraints), allowing three internal mobilities. Alternatively, the closures can be represented with 3D constraints.
\subsection{Kangaroo}
\begin{figure}
     \centering
     \begin{subfigure}[b]{0.45\textwidth}
         \centering
         \includegraphics[width=0.32\textwidth]{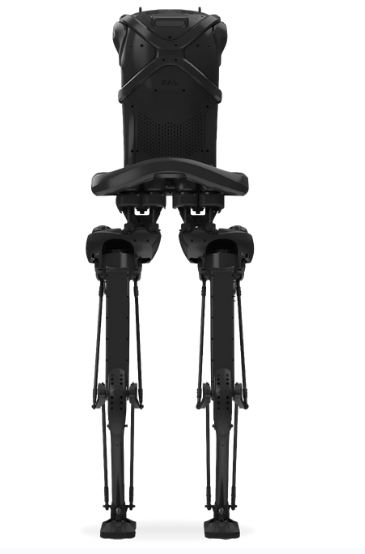}
         \caption{Kangaroo Robot}
         \label{fig:kangarooRobot}
     \end{subfigure}
     \hspace{0cm}
     \begin{subfigure}[b]{0.45\textwidth}
         \centering
         \includegraphics[width=0.32\textwidth]{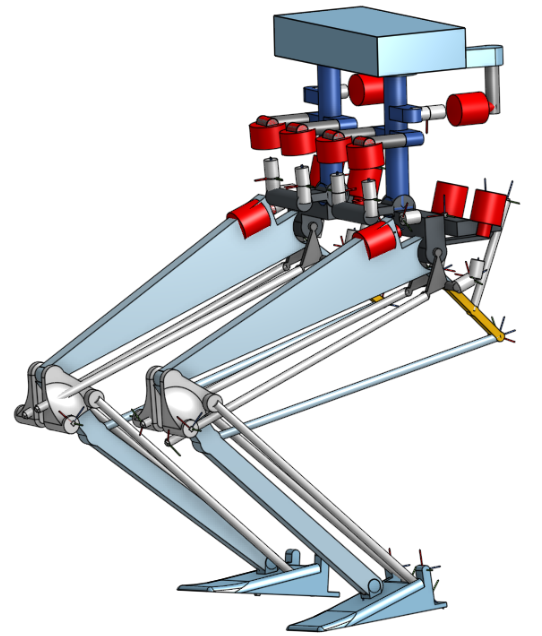}
         \caption{Modeled kangaroo robot}
         \label{fig:kangarooModel}
     \end{subfigure}
    \caption{Real and modeled robot. The model is presented with open 6D and 3d contact constraint}
    \label{fig:kangaroo}
\end{figure}

Kangaroo, a high-dynamic humanoid from Pal Robotics, features legs actuated by 6 linear motors. Its complexity surpasses Digit, as all actuators are in the hips, resulting in 12 kinematic closures (Fig.~\ref{fig:kangaroo}). 
The real and modeled robot are presented in figure \ref{fig:kangaroo}.
The model includes a mix of three 6D constraints and nine 3D constraints, yielding 45 constrained DoF within a 63 DoF serial model. This results in 12 internal mobilities, where each closure rod rotates freely around its axis.
Other parallel robot examples are provided inside the git repository.

%% file: tex/5.Conclusion.tex
\section{Conclusion} \label{sec:conclusion}

The code for generating and parsing the extended files, along with a collection of example models, can be found on GitHub\footnote{\href{https://github.com/Gepetto/example-parallel-robots}{https://github.com/Gepetto/example-parallel-robots}}. 
Additionally, we provide the code of several utility functions for working with parallel robots\footnote{\href{https://github.com/Gepetto/toolbox-parallel-robots}{https://github.com/Gepetto/toolbox-parallel-robots}}. 
These resources demonstrate the practical implementation of our proposed extension and serve as a foundation for further exploration and development for other specific software.
%

Our paper introduced an extension to URDF to support closed-loop kinematic structures, addressing the need for a description format for parallel robots. By overcoming the limitations of standard URDF, our approach enables more comprehensive and efficient modeling of complex robot topologies while maintaining compatibility with existing simulation and control frameworks. This extension integrates seamlessly into current workflows, simplifying the modeling process and enhancing URDF’s usability for advanced applications. Future work will focus on making this extension work on different software and simulators.